\documentclass[letterpaper, 10 pt, conference]{ieeeconf}  % Comment this line out if you need a4paper

\IEEEoverridecommandlockouts                              % This command is only needed if 
\usepackage{cite}
\usepackage{amsmath,amssymb,amsfonts}
\usepackage{algorithm}
\usepackage{algpseudocode}[noend]
\usepackage{subcaption}
\usepackage{graphicx}   
\usepackage{physics}
\usepackage{textcomp}
\usepackage{hyperref}
\usepackage{gensymb}
\usepackage{multirow}
\let\labelindent\relax
\usepackage{enumitem}
\usepackage[font=small,labelfont=bf]{caption}
\usepackage[table,xcdraw]{xcolor}
\usepackage{adjustbox}
\usepackage{dsfont}
\usepackage{booktabs}
\usepackage{siunitx}
\usepackage{subfiles}
\usepackage{amssymb}
\usepackage{xcolor}

\usepackage{pifont}

\newcommand{\cmark}{\textcolor[HTML]{009933}{\ding{52}}}  % Green checkmark
\newcommand{\xmark}{\textcolor[HTML]{CC0000}{\ding{56}}}  % Red cross

\def\secref#1{Sec.~\ref{#1}}

\def\figref#1{Fig.~\ref{#1}}
\def\tabref#1{Tab.~\ref{#1}}
\def\eqref#1{Eq.~(\ref{#1})}
\def\algref#1{Alg.~\ref{#1}}

\usepackage[abbreviations]{glossaries-extra}

\glssetcategoryattribute{abbreviation}{indexonlyfirst}{true}

\glssetcategoryattribute{abbreviation}{nohyperfirst}{true}

\newabbreviation{auroc}{AUROC}{Area Under the Receiver Operating Characteristic Curve}
\newabbreviation{accuracy}{Acc}{Accuracy}

\newabbreviation{bev}{BEV}{Bird`s Eye View}
\newabbreviation{cnn}{CNN}{Convolutional Neural Network}

\newabbreviation{slam}{SLAM}{Simultaneous Localization and Mapping}
\newabbreviation{sota}{SoTA}{state-of-the-art}
\newabbreviation{fov}{fov}{field-of-view}
\newabbreviation{iou}{IoU}{Intersection over Union}

\title{\LARGE \bf
TartanGround: A Large-Scale Dataset for Ground Robot Perception and Navigation
}

\author{Manthan Patel$^{1}$, Fan Yang$^{1}$, Yuheng Qiu$^{2}$, Cesar Cadena$^{1}$, Sebastian Scherer$^{2}$, \\Marco Hutter$^{1}$ and Wenshan Wang$^{2}$ % <-this % stops a space
\thanks{$^1$Robotic Systems Lab, ETH Zurich, Zurich, Switzerland
}
\thanks{$^2$Robotics Institute of Carnegie Mellon University, Pittsburgh, USA}
}
\begin{document}
\bstctlcite{IEEEexample:BSTcontrol}

\maketitle
\thispagestyle{empty}
\pagestyle{empty}

%%%%%%%%%%%%%%%%%%%%%%%%%%%%%%%%%%%%%%%%%%%%%%%%%%%%%%%%%%%%%%%%%%%%%%%%%%%%%%%%

%%%%%%%%%%%%%%%%%%%%%%%%%%%%%%%%%%%%%%%%%%%%%%%%%%%%%%%%%%%%%%%%%%%%%%%%%%%%%%%%
\maketitle
\thispagestyle{empty}
\pagestyle{empty}

%----------------------------------------------%
%                  Abstract    
%----------------------------------------------%
\begin{abstract}

We present TartanGround, a large-scale, multi-modal dataset to advance the perception and autonomy of ground robots operating in diverse environments. This dataset, collected in various photorealistic simulation environments includes multiple RGB stereo cameras for 360-degree coverage, along with depth, optical flow, stereo disparity, LiDAR point clouds, ground truth poses, semantic segmented images, and occupancy maps with semantic labels. Data is collected using an integrated automatic pipeline, which generates trajectories mimicking the motion patterns of various ground robot platforms, including wheeled and legged robots. We collect 878 trajectories across 63 environments, resulting in 1.44 million samples. Evaluations on occupancy prediction and SLAM tasks reveal that state-of-the-art methods trained on existing datasets struggle to generalize across diverse scenes. TartanGround can serve as a testbed for training and evaluation of a broad range of learning-based tasks, including occupancy prediction, SLAM, neural scene representation, perception-based navigation, and more, enabling advancements in robotic perception and autonomy towards achieving robust models generalizable to more diverse scenarios. The dataset and codebase are available on the webpage: \url{https://tartanair.org/tartanground}

% Ground robots operate in diverse environments, from structured urban areas to unstructured terrains like forests and industrial sites. However, existing datasets are limited in scale or environment diversity, hindering the development of robust perception and navigation systems. While tasks such as semantic occupancy prediction have found great applications in autonomous driving, lack of established benchmarks for mobile robots operating in off-road and complex terrains have limited its applicability for ground robots operating in diverse environments. To this end, in this work we introduce TartanGround, a large-scale, multi-modal simulation dataset. It features 60 photorealistic environments created in Unreal Engine, encompassing dynamic lighting, adverse weather, and seasonal changes. The dataset includes multiple RGB stereo cameras for 360-degree coverage, along with depth, semantic segmentation, ground truth poses, optical flow, stereo disparity, LiDAR point clouds, and semantic occupancy maps. It provides W trajectories totaling in Y million samples, making it one of the most comprehensive datasets for ground robots. TartanGround supports a wide range of tasks, including occupancy prediction, neural scene representation, SLAM, and semantic segmentation, enabling advancements in robotic perception and autonomy across diverse and challenging environments.

\end{abstract}

% Diverse motion trajectories
% Add what experiments and evaluations we d
% Refine 
% Ending training and evalutation

%----------------------------------------------%
%                  Introduction    
%----------------------------------------------%
\section{Introduction}

Ground robots are increasingly being used in a wide range of environments, from structured urban areas to unstructured terrains such as forests, farmlands, and construction sites. These robots serve various purposes, including autonomous delivery, agricultural automation, industrial inspection, search-and-rescue missions, and construction site monitoring. To improve their adaptability and generalizability in these diverse settings, data-driven methods have gained traction. These approaches tackle key tasks in perception and scene understanding, such as \gls{slam}, occupancy prediction, semantic segmentation, monocular depth estimation, etc, enabling robots to better interpret, interact, and navigate their surroundings.\par

In the domain of autonomous driving, large-scale datasets~\cite{kitti, waymo2020, cordts2016cityscapes, caesar2020nuscenes} have played a crucial role in advancing machine learning models for tasks such as object detection, occupancy prediction, and semantic segmentation. These datasets offer standardized benchmarks that allow for consistent evaluation and comparison of algorithms. However, there is a notable lack of similar datasets and benchmarks for mobile robots operating in a broader range of environments. This absence makes it challenging to develop and evaluate generalizable models that can perform reliably across various and complex settings.\par

 \begin{figure}[t]
    \centering
    \includegraphics[width=\columnwidth]{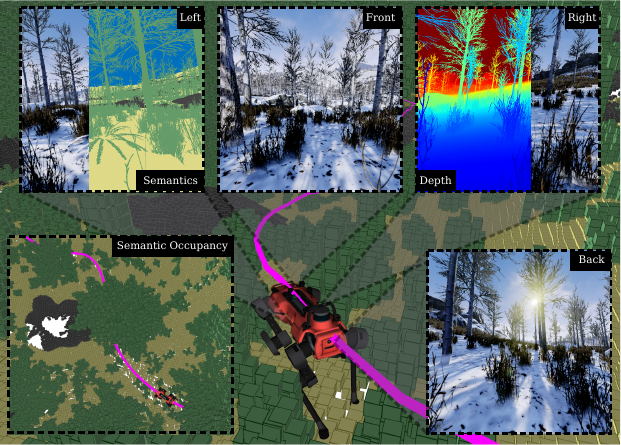}
    \caption{A trajectory from TartanGround (Winter Forest environment) includes multiple stereo RGB images covering a full 360$\degree$ \gls{fov}, along with accurate depth and semantic annotations. It also provides ground truth poses, LiDAR, IMU data, and semantic occupancy maps for comprehensive scene understanding.
    }
    \label{fig:cover_pic}
    \vspace{-2ex}
\end{figure}

Several datasets support research in mobile robotics, each tailored to specific environments. For example, datasets~\cite{jiang2021rellis, wigness2019rugd, mortimer2024goose, triest2022tartandrive, sivaprakasam2024tartandrive, vidanapathirana2024wildscenes} collected in off-road environments provide sensor modalities ranging from RGB cameras and LiDARs for scene understanding, to proprioceptive and traction data for vehicle dynamics modeling. Indoor datasets such as ScanNet~\cite{dai2017scannet}, TUM RGB-D~\cite{sturm12iros_tumrgbd}, and Matterport3D~\cite{chang2017matterport3d} provide \mbox{RGB-D} data for 3D reconstruction and \gls{slam}. However, these datasets often have limitations in environmental diversity, size, accurate ground truths, or sensor types, which limit their applicability in developing robust, generalizable models. \par

% Several datasets have been created to support research in mobile robotics, each focusing on specific areas. For outdoor settings, RELLIS-3D~\cite{jiang2021rellis} provides multi-modal data designed for off-road robotics, while RUGD~\cite{wigness2019rugd} offers semantic segmentation data for unstructured outdoor environments. WildScenes~\cite{vidanapathirana2024wildscenes} supplies synchronized image and LiDAR data with semantic annotations in natural settings, and TartanDrive~\cite{triest2022tartandrive, sivaprakasam2024tartandrive} offers extensive sensor data for learning dynamics models in off-road driving scenarios. Indoor datasets like ScanNet~\cite{dai2017scannet}, TUM-RGBD~\cite{sturm12iros_tumrgbd}, and Matterport-3D~\cite{chang2017matterport3d} provide RGB-D data for tasks such as 3D reconstruction and \gls{slam}. Despite their value, these datasets often have limitations in size, environmental diversity, or sensor types, which can limit their usefulness in developing robust, general models. \par

 \begin{figure*}[t]
    \centering
    \includegraphics[width=\textwidth]{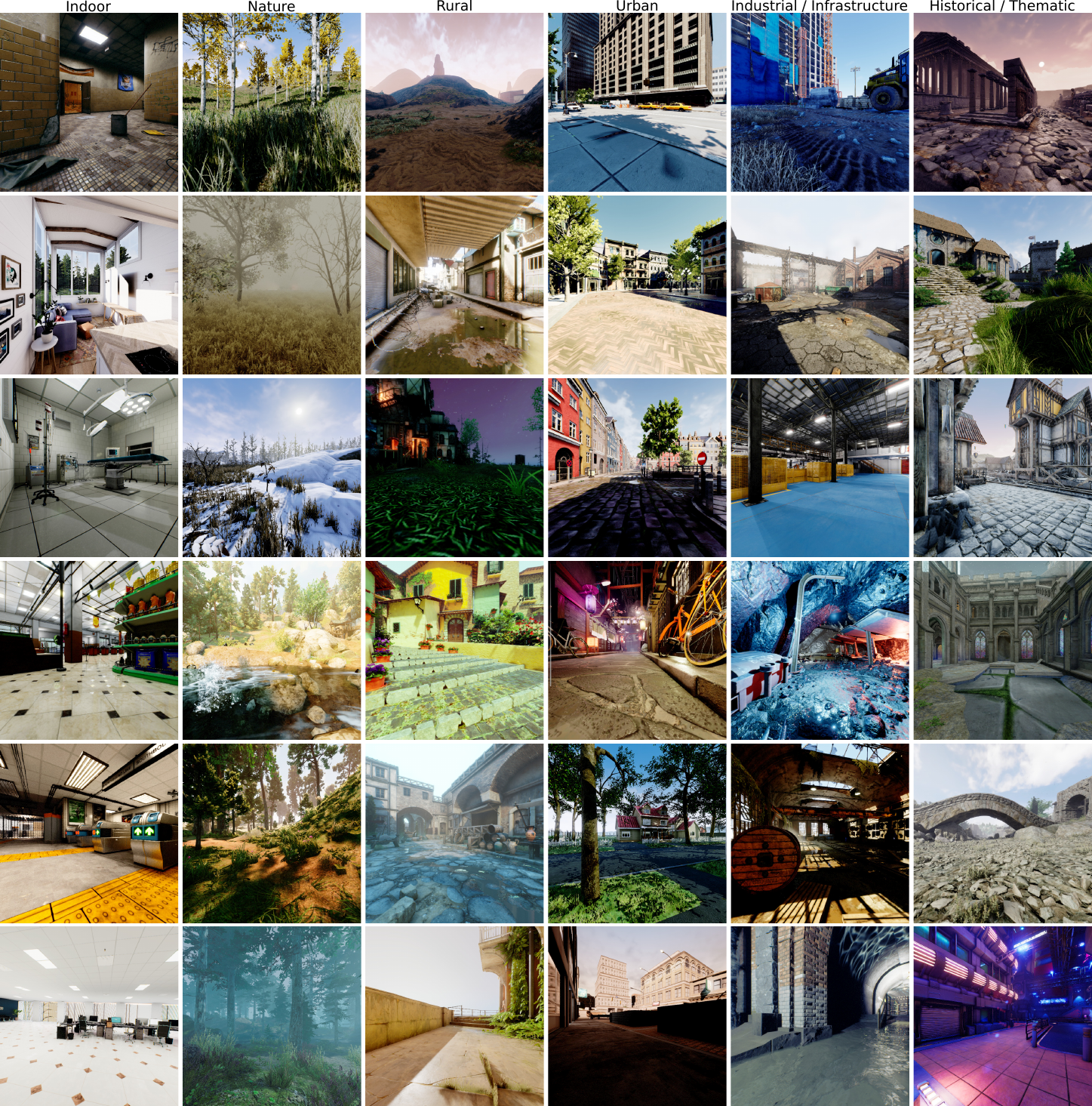}
    \caption{The TartanGround environments, categorized into Indoor, Nature, Rural, Urban, Industrial/Infrastructure, and Historical/Thematic
    }
    \label{fig:grid_env}
    \vspace{-2ex}
\end{figure*}

Simulation datasets have become an important alternative to overcome these difficulties and limitations of real-world data collection in robotics. They offer a controlled environment for generating large-scale, high-quality data with precise ground truth annotations for semantics, depth, and poses—information that is often difficult to obtain in real-world settings. Furthermore, simulation enables diverse data collection under varying lighting, weather, and terrain conditions, improving model robustness and generalization. These benefits have made simulation datasets crucial for advancing perception tasks such as \gls{slam} and monocular depth estimation. For instance, TartanAir~\cite{wang2020tartanair} has been instrumental in training \gls{sota} \gls{slam} algorithms like DROID-SLAM~\cite{teed2021droid}, demonstrating effective sim-to-real transfer. Similarly, synthetic datasets~\cite{ cabon2020virtual, wang2021irs, roberts2021hypersim} have played a key role in training foundation models for monocular depth estimation~\cite{yang2024depth, bochkovskii2024depth}, further showcasing the impact of learning from simulation data. \par

% The success of synthetic and simulation datasets has shown promise in overcoming these limitations. For example, TartanAir offers photorealistic simulation data that has been key in training advanced learning-based SLAM algorithms like DROID-SLAM, achieving SOTA results and impressive transferability from simulation to the real-world domain. Similarly, synthetic datasets such as ML Hypersim and Virtual KITTI (vKITTI) have been crucial in training foundational models for monocular depth estimation, as seen with Depth Anything V2 and ML-DepthPro. The benefits of simulation-based datasets include access to accurate ground truth for semantics, depth, and poses, the ability to generate large amounts of data, and the inclusion of a wide variety of environmental variations, all of which enhance model robustness and performance.

To address the need for comprehensive data resources, we introduce TartanGround, a large-scale simulation dataset designed to support perception and navigation tasks for ground robots across diverse environments. TartanGround includes data from over 63 diverse, challenging photorealistic environments, offering multi-modal sensor data such as multiple stereo RGB-D images covering a 360\degree \gls{fov}, semantic labels, LiDAR point clouds, semantic occupancy maps, and ground truth poses. We show that the existing \gls{sota} occupancy prediction methods, solely trained on data from the autonomous driving domain, do not generalize to environments of mobile robot operations such as forests. Moreover, we evaluate various \gls{sota} \gls{slam} algorithms and find that they struggle in challenging scenarios of low visibility and heavy occlusions. To summarize, the main contributions of this paper are: (1) A large dataset with over 1.44 million samples collected across diverse environments with precise ground truth labels mimicking different ground robot motion patterns, including wheeled (omnidirectional and differential-drive), and legged robots (quadrupedal), (2) an automatic data collection pipeline and (3) evaluation of two tasks, occupancy prediction and \gls{slam} highlighting the limitations of models trained on existing datasets.
% This dataset aims to set new benchmarks for tasks like semantic occupancy prediction, aiding in the development and assessment of algorithms in both structured and unstructured settings.

% Introduction of TartanGround: A comprehensive simulation dataset providing diverse environmental contexts and multi-modal sensor data to advance ground robot perception and navigation.

% Benchmark Establishment: Creation of standardized benchmarks for semantic occupancy prediction and other scene understanding tasks, enabling consistent evaluation of algorithm performance.

% Facilitation of Sim-to-Real Transfer: Demonstration of the dataset's effectiveness in training models that generalize well from simulation to real-world applications, bridging the gap between experimental research and practical use.

% With TartanGround, we aim to offer the robotics community a valuable resource that addresses current dataset limitations, promoting advancements in robotic autonomy across a wide range of environments.

%----------------------------------------------%
%                  Dataset Features
%----------------------------------------------%
\section{Related Work}

In the field of autonomous driving, the KITTI~\cite{kitti} dataset offers a comprehensive suite of sensor data, including stereo camera and LiDAR inputs, primarily for autonomous driving research. Building upon this, SemanticKITTI~\cite{behley2019semantickitti} extends KITTI by providing dense point-wise semantic labels for LiDAR scans, enabling advancements in LiDAR segmentation tasks. NuScenes~\cite{caesar2020nuscenes} and Waymo Open Dataset~\cite{waymo2020} further contribute to urban scene understanding by offering 360$\degree$ sensor coverage and diverse urban scenarios. Occ3D~\cite{tian2023occ3d} establishes new benchmarks for semantic occupancy prediction on nuScenes and Waymo, facilitating the development of models that predict both the geometry and semantics of urban environments.

In indoor environments, several RGB-D datasets have been instrumental in advancing scene understanding. ScanNet~\cite{dai2017scannet} comprises annotated 3D reconstructions of indoor scenes, serving as a benchmark for tasks like 3D semantic segmentation and object recognition. TUM RGB-D~\cite{sturm12iros_tumrgbd} offers sequences recorded with handheld cameras, providing ground truth trajectories for evaluating SLAM and odometry algorithms. SUN RGB-D~\cite{song2015sun} and NYUv2~\cite{silberman2012indoor}  provide paired RGB-D with semantic labels, supporting research in semantic segmentation and monocular depth estimation. Matterport3D~\cite{chang2017matterport3d} offers a rich collection of indoor images, facilitating research in 3D reconstruction and navigation.

For off-road environments, RELLIS-3D~\cite{jiang2021rellis} provides multi-modal data with dense annotations, supporting semantic segmentation research. RUGD~\cite{wigness2019rugd} offers images captured in unstructured outdoor environments with pixel-wise semantic labels. GOOSE~\cite{mortimer2024goose} presents data collected with a vehicle equipped with multiple cameras and LiDARs ensuring 360$\degree$ coverage in offroad environments. WildScenes~\cite{vidanapathirana2024wildscenes} provides synchronized image and LiDAR data with semantic annotations in natural settings, while TartanDrive~\cite{triest2022tartandrive} offers extensive sensor data for learning dynamics models in off-road driving scenarios. WildOcc~\cite{zhai2024wildocc} builds upon RELLIS-3D by providing semantic occupancy annotations, enabling research in 3D scene understanding in off-road context. 

Despite the significant contributions of these datasets, they are often limited by scale, environmental diversity, or sensor modalities, which can hinder the development of generalizable models for robotic perception and navigation. To address these limitations, TartanAir~\cite{wang2020tartanair} introduced a large-scale synthetic dataset with 20 diverse environments and random motion patterns, enhancing generalization by offering photo-realistic imagery under varied weather and lighting conditions. TartanAir-V2 extended V1 with more scenes and more modalities~\cite{tartanairv2}. Building on this, we introduce TartanGround, specifically targeting ground robots. It features realistic ground robot motion patterns, including wheeled and legged robots. We summarize the various datasets along with the sensor data, available ground truth, and scale in \tabref{tab:dataset_comparison}. TartanGround is the only large-scale dataset covering diverse environments and having realistic ground robot motions.

\begin{table*}[t]
\centering
\caption{Comparison of Datasets for Robotic Perception}
\resizebox{\textwidth}{!}{%
\begin{tabular}{llllccccccccc}
\hline
\textbf{Dataset} & \textbf{Type} & \textbf{Env} & \textbf{Motion Pattern} & \textbf{Stereo} & \textbf{Depth} & \textbf{LiDAR} & \textbf{Semantics} & \textbf{Occupancy} & \textbf{Pose} & \textbf{Multi-cam} & \textbf{Seq Num} & \textbf{Samples} \\ 
\hline
Semantic KITTI~\cite{behley2019semantickitti}  & Real & Urban   & Car              & \cmark & \xmark  & \cmark & \cmark & \cmark & \cmark & \xmark & 22   & 43 K   \\
Occ3D-Waymo~\cite{waymo2020, tian2023occ3d}     & Real & Urban   & Car             & \xmark & \xmark  & \cmark & \cmark & \cmark & \cmark & 5 & 1000    & 200 K   \\
Occ3D-nuScenes~\cite{caesar2020nuscenes, tian2023occ3d}  & Real & Urban   & Car    & \xmark & \xmark  & \cmark & \cmark & \cmark & \cmark & 6 & 1000    & 40 K   \\
RELLIS-3D~\cite{jiang2021rellis}       & Real & Off-road& Car                      & \cmark & \cmark  & \cmark & \cmark & \xmark & \cmark & \xmark & 5    & 6.2 K  \\
RUGD~\cite{wigness2019rugd}            & Real & Off-road& Car                      & \xmark & \xmark  & \xmark & \cmark & \xmark & \xmark & \xmark & 18   & 7.5 K  \\
TartanDrive~\cite{triest2022tartandrive}     & Real & Off-road& Car                & \cmark & \cmark  & \xmark & \xmark & \xmark & \cmark & \xmark & 630  & 184 K  \\
WildScenes~\cite{vidanapathirana2024wildscenes}      & Real & Off-road& Hand       & \xmark & \xmark  & \cmark & \cmark & \xmark & \cmark & \xmark & 5    & 9.6 K  \\
WildOcc~\cite{zhai2024wildocc}         & Real & Off-road& Car                      & \cmark & \cmark  & \cmark & \cmark & \cmark & \cmark & \xmark & 5    & 10 K   \\
SynPhoRest~\cite{nunes_2022_synphorest}         & Syn. & Off-road& Random                      & \xmark & \cmark  & \cmark & \cmark & \xmark & \xmark & \xmark & X    & 3.1 K   \\
GOOSE~\cite{mortimer2024goose}           & Real & Off-road& Car            & \xmark & \xmark  & \cmark & \cmark & \cmark & \cmark &  4    & 356   & 10 K\\
GOOSE-Ex~\cite{hagmanns2024excavating}    & Real & Off-road& Tracked, Legged            & \xmark & \xmark  & \cmark & \cmark & \cmark & \cmark &  4    & 100   & 5 K\\
Scannet~\cite{dai2017scannet}         & Real & Indoor  & Hand                      & \xmark & \cmark  & \xmark & \cmark & \cmark & \cmark & \xmark & 1500 & 2.5 M  \\
GrandTour~\cite{frey2025boxi}       & Real & Mix     & Legged, Wheeled             & \cmark & \cmark  & \cmark & \xmark & \xmark & \cmark & 3 & 100   & 100 K  \\
TartanAir~\cite{wang2020tartanair}       & Syn. & Mix     & Random                 & \cmark & \cmark  & \xmark & \cmark & \xmark & \cmark & \xmark  & 1037   & 400 k   \\
TartanAir-V2~\cite{tartanairv2}       & Syn. & Mix     & Random                 & \cmark & \cmark  & \cmark & \cmark & \cmark & \cmark & 6  & 1110   & 1.4 M   \\
TartanGround    & Syn. & Mix     & Legged, Wheeled                                 & \cmark & \cmark  & \cmark & \cmark & \cmark & \cmark & 6  & 878    & 1.4 M \\ 
\hline
\end{tabular}
}
\label{tab:dataset_comparison}
\end{table*}

%----------------------------------------------%
%               Problem Formulation    
%----------------------------------------------%
\section{The Dataset}
 \begin{figure*}[t]
    \centering
    \includegraphics[width=\textwidth]{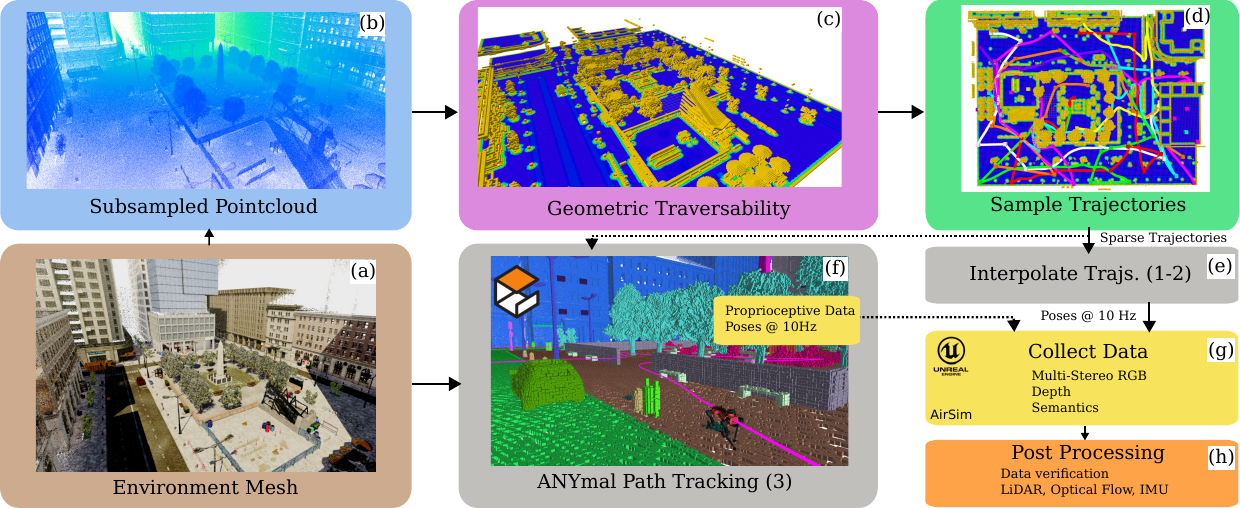}
        \caption{Overview of the data collection pipeline. We first subsample a pointcloud (b) from the environment mesh (a), which is used to generate the geometric traversability (c). Next, we sample sparse long trajectories covering the environment, which are either interpolated to dense poses for wheeled robots (e) or used in a gazebo simulation with path tracking for quadrupeds (f) to generate the dense poses. The photorealistic data is then collected in AirSim using the poses (g), followed by a post-processing step (h).}
    \label{fig:pipeline}
\end{figure*}

\subsection{Features}

TartanGround consists of 63 realistic simulation environments from TartanAir-V2~\cite{tartanairv2}, covering diverse scenarios from structured urban to large-scale unstructured outdoor environments~(\figref{fig:grid_env}). The environments have been developed with Unreal Engine 4, and the data is collected using the AirSim~\cite{shah2018airsim} plugin. This setup allows for rendering photorealistic scenes with high fidelity and makes it possible to have dynamic lighting, adverse weather effects, dynamic objects, and seasonal changes. In each of the environments, we collect data from multiple sampled trajectories with diverse motion patterns mimicking real-world ground robots. To achieve full 360$\degree$ coverage, we use 6 stereo RGB cameras (front, left, right, back, top, bottom), each with a \gls{fov} of 90$\degree$. Each camera is synchronized with accurate ground truth poses and also records depth and semantic segmentation images. During post-processing, we generate additional ground truth data, including IMU data, optical flow, stereo disparity, LiDAR point clouds, and semantic occupancy maps. In total, we collect 878 trajectories (440 omni-wheeled, 198 diff-wheeled, and 240 legged), each with 600 to 8000 samples, resulting in a total dataset size of approximately 15 TB, 1.44 million samples, and 17.3 million RGB images. This large-scale, multi-modal dataset is designed to support a wide range of robotic perception and navigation tasks, benefiting the community in establishing new benchmarks for generalizable learning-based methods.

\subsection{Trajectory Sampling}
The overview of the trajectory sampling pipeline is shown in \figref{fig:pipeline}.
\subsubsection{Environment Pointcloud}
We export the corresponding mesh~(\figref{fig:pipeline}a) from Unreal Engine for each environment and sample a point cloud~(\figref{fig:pipeline}b) with a specified density $\delta$. We exclude foliage elements such as grass and small bushes from the mesh export to ensure our geometry-based traversability estimation accurately reflects navigable paths. This approach allows us to generate trajectories that traverse compressible vegetation, capturing more realistic and diverse data.
% For each environment, we export the respective mesh from Unreal Engine and subsample a point cloud with density $\delta$. Here, we avoid exporting the foliage actors, such as grass and small bushes, into the mesh as our traversability estimation approach is based on geometry, and we would like to collect trajectories going over this compressible vegetation.
\subsubsection{Traversability Generation}
We utilize a geometry-based traversability estimation pipeline inspired by \cite{yang2024efficient}, which, taking as an input the environment point cloud, can efficiently represent the environment with complex terrain conditions and spatial structures into 3D tomogram slices~(\figref{fig:pipeline}c). This method extends traditional elevation maps~\cite{fankhauser2014robot} for multi-layered environments by assigning ground and ceiling elevations at fixed intervals, facilitating efficient planning of 3D trajectories for ground robots. Moreover, it incorporates the robot's capabilities, such as climbing stairs and navigating steep gradients, by assessing factors like slope, step height, and overhead clearance to compute the traversability values. This approach ensures robust performance in both structured and unstructured multi-layered environments.

% efficiently represents a multi-level environment by assigning ground and ceiling elevations at fixed intervals facilitating efficient planning of 3D trajectories for ground robots. Moreover, it allows for incorporating the robot's capabilities, such as climbing stairs and steep gradients, while generating scene traversability. 
% We employ a geometry-based traversability estimation pipeline inspired by \cite{yang2024efficient}. This approach takes as an input a point cloud of the environment, capturing intricate terrain features and spatial structures. To efficiently represent complex 3D terrains, we generate tomogram slices from this point cloud, encoding both ground and ceiling elevations. This tomographic representation extends traditional elevation maps, enabling the depiction of multi-layered environments while maintaining computational efficiency.
% The traversability analysis evaluates each tomogram slice, considering the robot's locomotion capabilities, such as its ability to navigate stairs or steep slopes. By assessing factors like slope, step height, and overhead clearance, the system determines the navigability of various terrain segments. The resulting traversability maps inform the planning module, facilitating the generation of safe and efficient 3D trajectories that account for the robot's physical constraints and the environment's complexity.
% By integrating this advanced traversability estimation pipeline, our system effectively navigates diverse and challenging terrains, ensuring robust performance in both structured and unstructured environments.

\subsubsection{Sparse Trajectory Sampling}
% Our objective is to sample $\mathcal{S}$ long trajectories in the environments such that they maximize the coverage of the environments. We uniformly sample $n/2$ points in free space and $n/2$ points near obstacles, ensuring that trajectories cover both open and constrained regions. To reduce redundancy and maintain spatial diversity, we apply a representative point sampling step using k-means clustering to obtain a set of $K$ representative points. Each of the representative point is then randomly assigned a subgroup $s \in S$. For each subgroup $\mathcal{S}$, we generate a graph where the nodes represent the points, and the edges represent the path distance between the points in the tomogram. The graph is then processed using a traveling salesman problem (TSP) approximation to determine an efficient traversal sequence. For each pair of consecutive nodes in the traversal sequence, we find the optimal smooth path using an A$\star$ approach for tomograms~\cite{yang2024efficient}. The final path is obtained by concatenating the subpaths between consecutive nodes ensuring connectivity. In the end we get $\mathcal{S}$ sparse trajectories which provide a good coverage of the environment.

Our objective is to generate $\mathcal{S}$ long trajectories that maximize coverage of the environment while maintaining spatial diversity~(\figref{fig:pipeline}d). To achieve this, we begin by uniformly sampling $n/2$ points in free space and $n/2$ points near obstacles, ensuring that the trajectories encompass both open and constrained regions. To eliminate redundancy, we apply a representative point sampling step using k-means clustering, resulting in a set of $K$ representative points. \par

Each representative point is randomly assigned to one of the $\mathcal{S}$ trajectory subgroups. For each subgroup, we construct a graph where nodes correspond to the sampled points and edges represent the path distance between them in the tomogram. We approximate the Traveling Salesman Problem (TSP) on this graph to determine an optimal traversal sequence. For each pair of consecutive nodes in the traversal sequence, we find the optimal smooth path using an A$\star$ approach for tomograms~\cite{yang2024efficient}. The final path is obtained by concatenating the subpaths between consecutive nodes, ensuring connectivity. This approach is summarized in \algref{alg:sparse_sampling}. By the end of this process, we obtain $\mathcal{S}$ sparse trajectories that provide comprehensive coverage of the environment.

\begin{algorithm}[h]
\caption{Sparse Trajectory Sampling}
\label{alg:sparse_sampling}
\begin{algorithmic}[1]
\State \textbf{Input:} Tomogram $\mathcal{T}$, samples $n$, trajectories $\mathcal{S}$
\State \textbf{Output:} Sparse trajectories $\mathcal{X} = \{X_1, X_2, ..., X_S\}$
\State $\mathcal{T}_{free} \gets$ SampleFreeSpace$(\mathcal{T}, n/2)$
\State $\mathcal{T}_{obs} \gets$ SampleNearObstacles$(\mathcal{T}, n/2)$
\State $\mathcal{T}_r \gets$ KMeans$(\mathcal{T}_{free} \cup \mathcal{T}_{obs}, K)$
\State $\mathcal{S} \gets$ RandomAssign$(\mathcal{T}_r, S)$

\For{$s \in \mathcal{S}$}
    \State $G_s = (V_s, E_s)$, where $V_s = \mathcal{T}_r^s$, $E_s =$ path distances
    \State $\pi_s \gets$ SolveTSP$(G_s)$
    \For{$(v_i, v_{i+1}) \in \pi_s$}
        \State $P_{i, i+1} \gets A^*(v_i, v_{i+1})$
        \State $X_s \gets \text{Concat}(X_s, P_{i, i+1})$
    \EndFor
    
\EndFor
\State \Return $\mathcal{X} = \{X_1, ..., X_S\}$

\end{algorithmic}
\end{algorithm}

\subsubsection{Dense Trajectory Generation}
% In order to collect the data with AirSim, we need dense poses at a fixed frequency (\SI{10}{Hz}) for which we need to interpolate the sparse trajectories in a realistic manner so as to mimic the motion of a ground robot with realistic velocity and acceleration constraints. For this, we employ three different variations, 1) omni-directional, 2) differential, and 3) legged robot trajectory (\secref{sec:legged_traj}). 

% path tracking with a fixed look ahead distance so that the heading of the robot is towards the point it is tracking. In omni directional we allow for any kind of motion in the ground plane, thus robot might be translating in a direction different to the heading, while for differential we restrict this kind of motion and thus it would turn in place to orient its heading to the next tracking point before starting to move in that direction. 

% simulate natural velocity variations by incorporating a random walk model for velocity changes while sampling acceleration values within predefined limits at each timestep. Additionally, smooth yaw transitions are applied using a bounded yaw rate, ensuring physically plausible motion profiles. 

% gaussian noise to the position 

To collect data in AirSim, we generate dense poses at a fixed frequency of \SI{10}{Hz}, requiring realistic interpolation of sparse trajectories to accurately mimic ground robot motion while adhering to velocity and acceleration constraints. We implement three trajectory variations: (1) omnidirectional, (2) differential-drive (\figref{fig:pipeline}e), and (3) legged robot trajectories (\secref{sec:legged_traj}).

We use a fixed look-ahead distance for path tracking, ensuring the robot moves toward its tracking point. In omnidirectional motion, the robot can translate in any direction on the ground plane, meaning its heading may not always align with its movement direction. In contrast, the differential-drive model enforces a stricter motion constraint, requiring the robot to reorient itself to the next tracking point before proceeding. We apply a random walk model to introduce realistic velocity variations, adjusting speeds dynamically while sampling acceleration values within predefined limits at each timestep. Additionally, we ensure smooth yaw transitions using a bounded yaw rate, maintaining physically plausible motion. We randomly sample the robot height in the range \( [0.5, 1.5] \, \text{m} \) for each trajectory. We also introduce Gaussian noise to the position to simulate real-world uncertainties, adding slight perturbations that account for terrain roughness and sensor and actuation inconsistencies.

\subsection{Legged Robot Trajectories}
\label{sec:legged_traj}

To capture realistic motion patterns of a legged robot, we perform path tracking of the sparse trajectories within a Gazebo simulation environment using an ANYmal D legged robot~(\figref{fig:pipeline}f). An example of the ANYmal robot in action in a forest environment is shown in \figref{fig:cover_pic}. \par
During simulation, we record the ground truth base poses, which are later used for sampling photorealistic data in AirSim. In addition to the base poses, we collect a comprehensive set of proprioceptive data, including base velocities and accelerations, joint states, and contact forces for each leg. In addition to the perception tasks, these trajectories can also be used for learning navigation tasks.

% To obtain motion patterns of a legged robot, we use path tracking of the sparse trajectory in a Gazebo simulation environment of select environments for an ANYmal D legged robot. For these environments, we import the environment mesh in Gazebo simulation, and do path tracking
% realistic physics and has shown effective verification for sim-to-real policies. A sample can be seen in \figref{fig:cover_pic}. We record the groundtruth base poses which can then be used for sampling the photorealistic data in airsim. We also record other proprioceptive data of the robot which consists of the base position, velocities and accelerations along with the joint states and the contact forces for each leg. 

\subsection{Data Collection, Verification and Post-processing}
For the provided dense poses~(\figref{fig:pipeline}e-f), we capture 6 RGB stereo image pairs in AirSim, along with corresponding depth and semantic segmentation images. Using an approach similar to TartanAir~\cite{wang2020tartanair}, we generate additional ground truth data, including optical flow, stereo disparity, and simulated LiDAR measurements from these raw images. To enhance the usability of the dataset, we introduce a custom camera resampling feature that enables image extraction with arbitrary intrinsics and rotation matrices. This allows users to specify camera parameters that match their real robot setup, and the system re-renders images accordingly from the captured set of 6 images, ensuring compatibility with diverse robotic platforms.

% For the provided dense poses, we capture 6 RGB stereo image pairs from the front, left, right, back, top, and bottom cameras in AirSim. The corresponding depth and semantic segmentation images are also recorded. Using an approach similar to TartanAir~\cite{wang2020tartanair}, we generate additional ground truth data, including optical flow, stereo disparity, and simulated LiDAR measurements from these raw images. To enhance the usability of data collection, we introduce a new custom camera resampling feature, which allows for extracting images with any arbitrary intrinsics and rotation matrices. Thus, the user can specify the parameters according to their real robot setup and we render the images accordingly from our captured set of 6 images.

% dynamic adjustment of camera intrinsics and orientations.
% This feature enables us to simulate a wide range of camera setups, enhancing the dataset's diversity and applicability across various robotic perception tasks.

For data verification, we ensure the synchronization between camera poses and captured images by computing the optical flow between consecutive image pairs and evaluating the mean photometric error, as described in TartanAir~\cite{wang2020tartanair}. Additionally, depth images are analyzed to verify collision occurrences with the environment.

% % TODO: Wenshan -> Add two lines on how exactly the custom camera resampling feature works
% Define custom cameras ->allows for camera resampling with specified intrinsics and rotation -> 

%----------------------------------------------%
%                 Experiments
%----------------------------------------------%
\section{Experiments and Applications}

% The TartanGround dataset can be used for various perception and navigation tasks. 
In this section,  we evaluate the performance of state-of-the-art methods on two key tasks: Occupancy Prediction and SLAM. We further discuss the other potential applications of the dataset.

\subsection{Occupancy Prediction}
The task of predicting 3D occupancy voxels using multi-camera images has become quite popular in the field of autonomous driving as it enables detailed spatial representation of the environment useful for downstream tasks such as path planning and navigation. End-to-end occupancy prediction networks~\cite{tian2023occ3d} have the advantage of handling occlusions and satisfying multi-camera consistency where traditional methods struggle. NuScenes~\cite{caesar2020nuscenes} and Waymo~\cite{waymo2020} have become popular benchmarks for evaluating these learned networks, however, there is a lack of similar benchmarks for mobile robots operating in diverse environments. We show through experiments that the methods trained on autonomous driving data do not generalize to other environments, and thus, there is a need for such a large-scale dataset for training and evaluating in different scenes. 

\subsubsection{Setup, Baselines, and Environments}

We set up two baselines for occupancy prediction. The first is a simple baseline that uses an off-the-shelf monocular depth estimator~\cite{bochkovskii2024depth} to project pixels into 3D space using the predicted depth. To reduce the effect of bleeding artifacts, we further apply gradient filtering. This baseline is designed to highlight the limitations of depth-based projections and emphasize the need for dedicated occupancy prediction networks. Our second baseline is SurroundOcc~\cite{wei2023surroundocc}, a state-of-the-art 3D occupancy prediction network. This network takes multiple RGB images as input and extracts per-image multi-scale features. Using 2D-3D spatial attention, the multi-camera information is fused to construct 3D multi-scale feature volumes which are decoded into semantic occupancy predictions. \par

\begin{figure}[]
    \centering
    \includegraphics[width=\columnwidth]{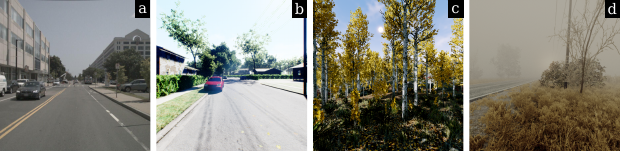}
    \caption{(a) nuScenes urban environment (training data), (b) ModNeighborhood, (c) ForestAutumn, and (d) GreatMarsh, from TartanGround which exhibit increasing difference in distributions compared to the nuScenes.}
    \label{fig:occ_env}
    \vspace{-2ex}
\end{figure}

For evaluating the baselines, we select three trajectories each from urban and natural environments, and truncate them to 1000 samples per trajectory. The urban environments depict city-like scenarios that are closer to the training domain of scenes (\figref{fig:occ_env}), while the natural environments include forest environments in different seasons and a marsh environment with heavy fog and low visibility. SurroundOcc network was trained on nuScenes data which was collected using a car mounted with six cameras having overlapping \gls{fov} and facing front-left, front, front-right, back-left, back and back-right directions at a resolution of 1600x900 pixels. To minimize the domain gap, we re-render images matching this setup using our image resampling pipeline. \par

We use the \gls{iou} metric to evaluate the performance of occupancy prediction. In principle, the network also predicts the semantic class along with the occupancy, however, since the labels of nuScenes do not match our environment labels, we do not evaluate this. Moreover, instead of evaluating in the range of $\pm$\SI{50}{m} from the egocentric frame in x-y directions as in SurroundOcc, we only evaluate in the range of $\pm$\SI{25}{m} (at \SI{0.5}{m} resolution) due to the relatively lower height of the cameras in our case limiting the longer distance visibility.

\subsubsection{Evaluations}

\begin{figure}[t]
    \centering
    \includegraphics[width=\columnwidth]{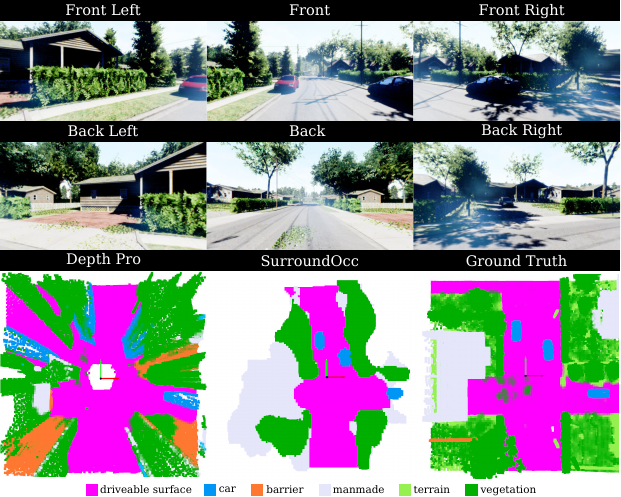}
    \caption{Qualitative results for the Occupancy Prediction task from the ModNeighborhood environment. The Y-axis (green) points towards the front of the robot. The segmentation colors are for visualization purposes only. For the Depth-Pro method, segmentation is obtained using SAN~\cite{xu2023side} with the nuScenes labels as vocabulary.
    }
    \label{fig:occupancy_pred}
    \vspace{-2ex}
\end{figure}

The quantitative results are summarized in \tabref{tab:occupancy_pred}. In general, we observe that SurroundOcc outperforms the depth-based projection pipeline across all urban environments. This is expected since urban environments have a similar distribution as the nuScenes training dataset. Qualitative results for the ModNeighborhood environment are shown in \figref{fig:occupancy_pred}. Here we see that SurroundOcc is able to predict the occupancy as well as the semantic classes (which we do not evaluate) for the car, vegetation, and driveable surface. On the other hand, visualizing the results of the depth projection method, we can clearly see the effect of bleeding and the inability to handle occlusions. Moreover, we also observe that this method suffers from inconsistent predictions across multiple views, leading to duplicating of objects when projected into 3D. These limitations highlight the advantages of occupancy prediction networks. 

The natural environments are completely out-of-distribution for SurroundOcc, and thus, the performance in these environments is significantly lower than the urban environments. Moreover, these environments are also more challenging, with lower visibility and higher occlusion due to the presence of dense trees and tall grass. Interestingly, here, the depth projection method performs much better highlighting the generalization capabilities of the monocular depth-estimator network. We believe that our large-scale dataset can contribute towards the development of robust generalizable models for predicting sematic occupancy in diverse environments.

% Nordic Harbor is a town with bridges and narrow streets surrounded by water. Since water was not part of the training data, fails to predict for this environment hence the low performance. 

\begin{table}[]
\centering
\renewcommand{\arraystretch}{1.2}
\caption{Occupancy Prediction IoU ($\uparrow$)}
\label{tab:occupancy_pred}
\resizebox{\columnwidth}{!}{%
\begin{tabular}{llcc}
\hline
Environment Name    & Type    & {DepthPro~\cite{bochkovskii2024depth}} & {SurroundOcc~\cite{wei2023surroundocc}} \\                    \hline
ModNeighborhood & Urban   & 17.29                              & \bf{19.67} \\
OldtownSummer       & Urban   & 17.37                              & \bf{22.15} \\ 
NordicHarbor        & Urban   & 11.62                              & \bf{12.91} \\ \hline
ForestAutumn        & Natural & \bf{16.59}                         & 13.89  \\
ForestWinter        & Natural & \bf{16.97}                         & 7.61   \\
GreatMarsh          & Natural & \bf{12.04}                         & 6.39
\\ \hline
\end{tabular}
}
\vspace{-2ex}
\end{table}

\subsection{Visual Odometry and SLAM}
\begin{figure}[h]
    \centering
    \includegraphics[width=\columnwidth]{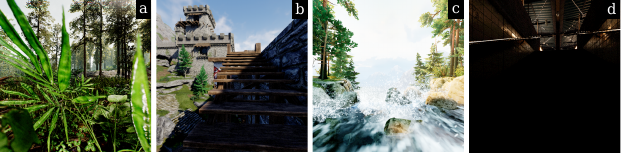}
    \caption{The environments used for testing SLAM, (a) Forest (heavy occlusion), (b) CastleFortress (indoor-outdoor transition), (c) WaterMillDay (running stream), (d) ModUrbanCity (dark stairs)
    }
    \label{fig:slam_env}
\end{figure}

Visual Odometry (VO) and SLAM for ground robots pose unique challenges, such as vegetation occlusions, short horizon caused by low altitude, and aggressive motion due to bumpy terrain. TartanGround is designed to capture these challenging cases (as shown in Fig.~\ref{fig:slam_env}). As a result, it becomes an interesting benchmark for VO and SLAM research. 

\subsubsection{Testing Environments and Baselines}
In this section, we show a small-scale evaluation of three state-of-the-art VO/SLAM algorithms on four challenging trajectories from four different environments. We use the front camera for these experiments. The trajectory from the Forest environment contains occlusion from tall grass and bushes, which are commonly presented in real-world off-road scenes. CastleFortress is a large scene with indoor-outdoor transitions, bringing dramatic illumination change. WaterMillDay has a rushing creek with water splash. ModUrbanCity contains a narrow dark stair that lacks good visual features. Our baseline consists of a classic geometry-based SLAM algorithm ORB-SLAM3~\cite{campos2021orb}, a learning-based monocular odometry DPVO~\cite{lipson2024deep}, and a stereo VO model MACVO~\cite{qiu2024mac}. We use relative translation error ($t_{\mathrm{rel}}$, $\mathrm{m/frame}$) and relative rotation error ($r_{\mathrm{rel}}$, $\mathrm{{}^\circ/frame}$) to evaluate the results. 

\subsubsection{Evaluation} 
Table~\ref{tab:slam} presents the quantitative results of various VO/SLAM algorithms on the TartanGround trajectories. To ensure a fair comparison, we interpolate missing data points and concatenate successfully tracked segments when algorithms lose tracking. Among the evaluated algorithms, ORB-SLAM3 frequently loses tracking, leading to significantly elevated translation and rotation errors. In contrast, MACVO and DPVO exhibit robustness across most environments. Notably, DPVO’s multi-frame design makes it more accurate in orientation estimation. However, its performance degrades in the presence of visual occlusions. MACVO is a frame-to-frame stereo odometry. It suffers from sudden viewpoint shifts but still shows exceptional translation accuracy. 
% Different from the existing SLAM benchmark which contains clear field-of-views.
TartanGround poses a unique challenge for the existing SLAM algorithms, and becomes a good complement to the existing SLAM benchmarks.
% The TartanGround benchmark provides realistic testing scenarios and accurate ground truth labels for ground robots.

% which are missing in existing SLAM benchmarks. 

% ORB-SLAM3 lost track of sequences \textcolor{red}{1 and 2. DPVO and MACVO are more robust to these challenging cases. DPVO performs better in occlusion cases because it has multi-frame optimization. MACVO suffers from heavy occlusion in the Forest scene but performs more robustly in other scenes}. 

\begin{table}[]
\caption{Performance comparison on the TartanGround Dataset. }
\renewcommand{\arraystretch}{1.2}
% \resizebox{\linewidth}{!}{
\begin{tabular}{lcccccc}
\toprule
\multirow{2}{*}{Scenes} & \multicolumn{2}{c}{ORB-SLAM3~\cite{campos2021orb}} & \multicolumn{2}{c}{DPVO$^{\star\dagger}$~\cite{lipson2024deep}} & \multicolumn{2}{c}{{MAC-VO}~\cite{qiu2024mac} } \\
\cmidrule{2-7}
       & $t_{\mathrm{rel}}$ & $r_{\mathrm{rel}}$ & $t_{\mathrm{rel}}$ & $r_{\mathrm{rel}}$ & $t_{\mathrm{rel}}$ & $r_{\mathrm{rel}}$ \\
\midrule
    Forest          &     0.152       &     1.667       &     0.010       &      0.039      &     \textbf{0.008}       &         \textbf{0.038}         \\
 CastleFortress  &        0.338    & 1.943 & 0.031  & \textbf{0.041} & \textbf{0.014} &          0.070       \\
    WaterMill     & 0.404 &      1.090      & 0.089    & \textbf{0.036}    & \textbf{0.016}    & 0.062 \\
    ModUrban     & 0.563 &       2.742     & 0.030    & \textbf{0.056}    & \textbf{0.011}    & 0.099 \\
\midrule
    \textbf{Average} & 0.364 & 1.861 & 0.040 & \textbf{0.043} & \textbf{0.012} & 0.067 \\
\bottomrule
\multicolumn{7}{l}{
$^\dagger$ Monocular method.
\quad
$^\star$ Scale-aligned with ground truth.} \\
\end{tabular}
% }
% \resizebox{\columnwidth}{!}{%
% \begin{tabular}{lllll}
% \hline
% Scene & Forest & CastleFortress & WaterMill & ModUrban  \\ \hline
% ORB-SLAM3~\cite{campos2021orb}   & -      & -              & -            & -            \\ 
% DPVO~\cite{lipson2024deep}  & -      & -              & -            & -      \\ 
% MACVO~\cite{qiu2024mac} & -      & -              & -            & -          \\ \hline
% \end{tabular}
% }

\label{tab:slam}
\end{table}

\subsection{Bird's Eye View Prediction}

\gls{bev} is an efficient way of representing the environments in the top-down view. This representation provides a comprehensive spatial understanding, facilitates the fusion of various sensor modalities, and is easily adaptable for downstream tasks such as object tracking and planning. Networks predicting semantic \gls{bev} maps in structured urban environments~\cite{philion2020lift, liu2023bevfusion} and predicting elevation and traversability \gls{bev} maps in unstructured environments~\cite{patel2024roadrunner, frey2024roadrunner,meng2023terrainnet} have gained popularity in recent years. TartanGround, with its diversity and scale, provides an ideal platform to advance these approaches.

\subsection{Neural Scene Representation}

\begin{figure}[t]
    \centering
    \includegraphics[width=\columnwidth]{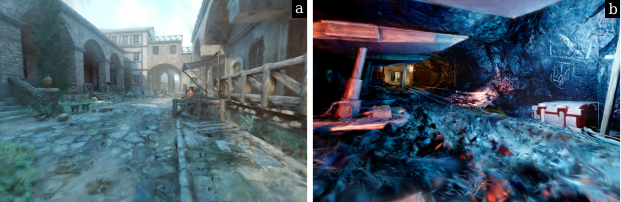}
    \caption{Novel view synthesis using Gaussian Splatting, trained on the Rome (a) and CoalMine (b) environments.
    }
    \label{fig:gaussian_splatting}
    \vspace{-2ex}
\end{figure}

The photorealism of our dataset makes it well-suited for advancing research in neural scene representation techniques, such as Gaussian splatting~\cite{kerbl20233d} and NeRFs~\cite{mildenhall2021nerf}, as well as neural SLAM methods~\cite{zhu2022nice, zhang2024glorie, keetha2024splatam, sandstrom2025splat}. Its large-scale scenarios, with dynamic lighting and adverse weather conditions, offer challenging test cases for novel view synthesis and robust scene reconstruction. As shown in \figref{fig:gaussian_splatting}, novel view synthesis using Gaussian splatting trained on trajectories from the Rome and CoalMine environments demonstrates the potential of TartanGround for high-fidelity neural rendering.

\subsection{Navigation}

The dataset can be used for various advanced navigation techniques in ground robotics. Recent research has demonstrated the potential of imitation learning and diffusion-based approaches for robust navigation in complex environments. Models such as NoMaD~\cite{sridhar2024nomad}, iPlanner~\cite{Yang-RSS-23}, and ViPlanner~\cite{roth2024viplanner} rely on diverse and large-scale datasets for training and evaluation. The TartanGround dataset, with its variety of environments, provides an ideal platform to develop and test these navigation models, enabling the creation of more robust and generalizable navigation systems for ground robots.
    
% The dataset can be leveraged for various advanced navigation techniques in ground robotics. Recent research has demonstrated the potential of imitation learning and diffusion-based approaches for robust navigation in complex environments. For instance, the NoMaD (Goal Masking Diffusion Policies for Navigation and Exploration) model utilizes a diffusion policy to handle both goal-directed navigation and goal-agnostic exploration, enabling effective navigation in unseen environments~\cite{sridhar2024nomad}. Similarly, iPlanner~\cite{Yang-RSS-23} and ViPlanner~\cite{roth2024viplanner} leveraged imperative learning that learns to map raw depth and RGB images to actions directly, enabling efficient navigation in unseen environments, by using data collected from simulation and real-world environments. The TartanGround dataset can be used to train and evaluate these navigation models in diverse environments, enabling the development of more robust and generalizable navigation systems for ground robots.

%----------------------------------------------%
%              Conclusion and Future Work
%----------------------------------------------%
\section{Conclusion and Future Work}

In this paper, we introduce TartanGround, a large-scale, multi-modal dataset designed to advance perception and navigation for ground robots in diverse environments. Our evaluations reveal the limitations of existing methods when applied to complex, unstructured environments, emphasizing the need for more robust and generalizable models. By providing comprehensive sensory data including multiple RGB stereo images, depth, semantic labels, LiDAR point clouds, semantic occupancy maps, and ground truth pose, TartanGround provides an ideal platform for training and benchmarking novel methods in occupancy prediction, SLAM, neural scene representation, visual navigation and more. In future, we aim to curate and release task-specific benchmarks utilizing the TartanGround dataset.

\section*{Acknowledgements}
This work was supported by the Luxembourg National Research Fund (Ref. 18990533) and the Swiss National Science Foundation (Ref. 200021E\_229503). This work used Bridges-2 at PSC through allocation cis220039p from the Advanced Cyberinfrastructure Coordination Ecosystem: Services \& Support (ACCESS) program which is supported by NSF grants 2138259, 2138286, 2138307, 2137603, and 2138296. The authors would like to thank Jannick Schröer for generating the Gaussian splatting renderings.

\bibliographystyle{IEEEtran}
\bibliography{ref}

\end{document}